\pdfoutput=1
\PassOptionsToPackage{dvipsames}{xcolor}
\PassOptionsToPackage{usenames, dvipsnames}{color}
\documentclass[11pt]{article}
\usepackage[preprint]{acl}

\usepackage{times}
\usepackage{latexsym}

\usepackage[T1]{fontenc}

\usepackage[utf8]{inputenc}

\usepackage{microtype}

\usepackage{inconsolata}

\usepackage{tabularx,colortbl}
\usepackage{hyperref}
\usepackage{amsmath}
\usepackage{amssymb}
\usepackage{algorithm}
\usepackage{algpseudocode}
\usepackage{graphicx}
\usepackage{multirow}
\usepackage{makecell} 
\usepackage{float}
\usepackage{caption}
\usepackage{subcaption}
\usepackage{soul}
\usepackage{comment}
\usepackage{arydshln}
\usepackage{hhline}
\usepackage{multicol}
\usepackage[normalem]{ulem}
\usepackage{bm}
\usepackage{makecell}
\usepackage{pbox}
\usepackage{tabularx}
\usepackage{MnSymbol}

\usepackage{lipsum}
\usepackage{fancyhdr}
\definecolor{My_color1}{rgb}{0.0, 0.66, 0.42}
\definecolor{My_color2}{rgb}{0.84, 0.23, 0.24}

\title{Do LLMs Know about Hallucination? An Empirical Investigation of LLM’s Hidden States}

\author{Hanyu Duan \,\,\,\,\,\,\,\,\,\, Yi Yang \,\,\,\,\,\,\,\,\,\, Kar Yan Tam \\
  Department of Information Systems, Business Statistics and Operations Management \\
  The Hong Kong University of Science and Technology \\
  \texttt{hduanac@connect.ust.hk} \,\,\,\,\,\,\,\,\,\, \texttt{\{imyiyang, kytam\}@ust.hk}}

\begin{document}
\maketitle
\begin{abstract}
Large Language Models (LLMs) can make up answers that are not real, and this is known as \textit{hallucination}. This research aims to see if, how, and to what extent LLMs are aware of hallucination. More specifically, we check whether and how an LLM reacts differently in its hidden states when it answers a question right versus when it hallucinates. To do this, we introduce an experimental framework which allows examining LLM’s hidden states in different hallucination situations. Building upon this framework, we conduct a series of experiments with language models in the LLaMA family \citep{touvron2023llama}. Our empirical findings suggest that LLMs react differently when processing a genuine response versus a fabricated one. We then apply various model interpretation techniques to help understand and explain the findings better. Moreover, informed by the empirical observations, we show great potential of using the guidance derived from LLM’s hidden representation space to mitigate hallucination. We believe this work provides insights into how LLMs produce hallucinated answers and how to make them occur less often.
\end{abstract}

\section{Introduction}

\begin{figure*}[!h]
\centering
\includegraphics[width=\linewidth]{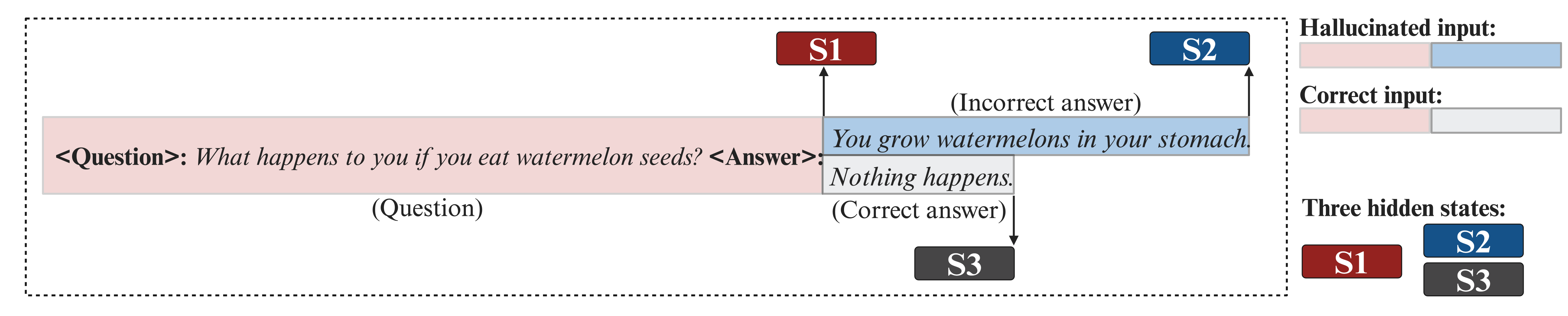}
\caption{Experimental framework, including two independent inputs (i.e., hallucinated input and correct input) and three critical hidden states ($\bm{s_1}$, $\bm{s_2}$, and $\bm{s_3}$). \textbf{<Question>} and \textbf{<Answer>} are templates that are adaptable and can be customized to suit various tasks.}
\label{fig:illustration}
\end{figure*}

Large language models (LLMs) like ChatGPT \footnote{\url{https://openai.com/chatgpt}}, GPT-4 \citep{openai2023gpt4}, PaLM \citep{chowdhery2022palm}, and LLaMA \citep{touvron2023llama}, have significantly changed the landscape of natural language processing (NLP), presenting great potential for advancing artificial general intelligence \citep{bubeck2023sparks}. However, the issue of \textit{hallucination} has simultaneously raised significant concerns \citep{rawte2023troubling, huang2023survey}. In the field of NLP, hallucination typically refers to a phenomenon where the generated content from LLMs is nonsensical, unfaithful, or incorrect in response to a given query \citep{filippova2020controlled, maynez2020faithfulness}. Recently, this issue has triggered a notable discussion among thousands of AI researchers around the world, resulting in over 30,000 signatures on an open letter \footnote{\url{https://futureoflife.org/open-letter/pause-giant-ai-experiments/}}, calling for a six-month pause on "giant AI experiment" \citep{pause2023experiments}. 

A growing body of NLP literature has examined hallucinations in LLMs, such as looking into the source of hallucinations from a training data perspective \citep{lin2021truthfulqa}, automating factuality detection for real-world claims \citep{chen2023complex}, and implementing automated feedback mechanisms to correct LLM-generated content \citep{pan2023automatically}. However, a fundamental research question is: \textit{Do LLMs possess awareness of hallucination, and if so, to what extent and through what mechanisms can they perceive it?}, which still remains underexplored.

In this work, we investigate LLM hallucination by analyzing its hidden states. In an autoregressive model, the hidden state encodes wealthy information about an LLM’s perception on the given input. Thus, comparing the paired hidden states between a correct response and a hallucinated one conditioned on the same query can provide insights into if and how LLMs differentiate between them. To this end, we introduce an experimental framework (as illustrated in Figure \ref{fig:illustration}), which involves providing \textbf{\textit{two}} inputs to the LLM and extracting \textbf{\textit{three}} critical hidden states to be analyzed. One input consists of a question followed an incorrect answer (e.g., \textit{Question: What happens to you if you eat watermelon seeds? Answer: You grow watermelons in your stomach}.), denoted as "\textbf{\textit{hallucinated input}}". The other input comprises the same question but followed by a correct response (e.g., \textit{Question: What happens to you if you eat watermelon seeds? Answer: Nothing happens}.), denoted as "\textbf{\textit{correct input}}". Passing the two inputs through the LLM individually, we extract three hidden states ($\bm{s_1}$, $\bm{s_2}$, and $\bm{s_3}$). $\bm{s_1}$ represents the final hidden state \footnote{Throughout this paper, we refer to "final hidden state" as the hidden state of the last input token in the last transformer layer.} of the question segment. Therefore, it encodes the LLM’s comprehension of the query and its prediction of a potential answer to the question. $\bm{s_2}$ relates to the final hidden state of the entire hallucinated input, containing the LLM’s perception after being exposed to the incorrect response. The same applied to $\bm{s_3}$. Thus, analyzing the transition of hidden states from $\bm{s_1}$ to $\bm{s_2}$ can provide insights into how a hallucinated answer alters the LLM’s final hidden state. Similarly, tracking the change between $\bm{s_1}$ and $\bm{s_3}$ can help understand how a correct answer influences the LLM’s final hidden state. Since the questions in both inputs are identical, that renders their hidden states comparable, which helps in understanding if and how the hidden states react differently when exposed to a correct answer versus a hallucinated one. 

We conduct experiments upon this framework (Figure \ref{fig:illustration}). We select LLaMA-2 7B, LLaMA-2-Chat 7B, and LLaMA-2 13B as the foundation LLMs \citep{touvron2023llama}. The experiments are performed on two datasets, TruthfulQA \citep{lin2021truthfulqa} and HaluEval \citep{li2023halueval}, with each sample consisting of a question paired with two answers, one being correct and the other hallucinated. By manipulating the inputs and employing various interpretation methods, our analysis of the hidden states reveals that LLMs are aware of hallucination and exhibit discernible reactions when provided with a correct answer compared to a hallucinated one. Particularly, we observe that the final hidden state is much more susceptible to the influence of a correct response compared to a hallucinated one. Based on this finding, we derive a measure that quantifies the extent to which an LLM possesses awareness of hallucination and find that strategically inducing an LLM to hallucination or including reference knowledge in the input can raise its awareness consequently, and the level of awareness corresponds to the uncertainty of the LLM in the responses. Moreover, we demonstrate the transition between hidden states encodes rich truthfulness information, that aligns closely with our awareness metric and shows great potential for hallucination mitigation (Section \ref{sec:case_study}). Additionally, from an information flow perspective \citep{wang2023label}, we observe that directly acquiring information from the question component of the input is crucial for preventing the generation of hallucinated answers, and the information within middle transformer layers proves more effective in detecting hallucinations compared to other layers.

In summary, this work makes two contributions. First, we provide empirical evidence indicating LLMs’ awareness of hallucination by showing disparities in the LLM's hidden representation space when processing accurate answers versus hallucinated ones. Second, the findings of this study shed light on mitigating LLM hallucination, potentially advancing the safe integration of foundation LLMs into critical downstream applications. We will release the experimental code for replication.

\section{Related Work}
According to the LLM hallucination taxonomy suggested by \citet{huang2023survey}, we position our work within three streams of relevant studies: 

\textbf{Hallucination causes.} Many works have focused on understanding hallucination causes from distinct perspectives, such as the training data \citep{kang2023impact}, the finetuning methods \citep{zhang2023language}, and the inference strategies \citep{chen2022towards}. Our study contributes to the inference line of research; however, unlike most existing works focused on analyzing the output vocabulary distribution \citep{chen2022towards}, ours explores the LLM’s internal representation space, opening up exciting new possibilities for understanding LLM hallucination.

\textbf{Hallucination detection.} Our work also relates to hallucination detection \citep{azaria2023internal, chern2023factool, min2023factscore}. In contrast to previous detection methods that often require using external knowledge sources for verification, the hallucinated cues presented in this study are derived solely from the LLM’s hidden representation space. A study closely related to ours, conducted by \citet{azaria2023internal}, similarly explores hidden states for hallucination detection; however, it directly utilizes hidden states as features and trains detection classifiers without delving into the mechanisms through which the hidden states encode truthfulness information, which is the primary focus of our work.

\textbf{Hallucination mitigation.} Our work also contributes to research in hallucination mitigation during the inference process \citep{pan2023automatically, dhuliawala2023chain, shi2023trusting}. Our study differs from existing efforts primarily by showing the feasibility of deriving guidance from the LLM’s hidden representation space and employing activation engineering techniques to mitigate hallucination. To the best of our knowledge, this study is the first to showcase the potential of utilizing activation engineering for mitigating LLM hallucination.

\section{Methodology}
In this section, we introduce our experimental framework, as illustrated in Figure \ref{fig:illustration}. 

At a higher level, the framework involves passing \textbf{\textit{two}} inputs through the LLM and obtaining \textbf{\textit{three}} critical hidden states accordingly. One input consists of a question along with its matching correct answer. We denote this input as "\textbf{\textit{correct input}}". (e.g., \textit{Question: What happens to you if you eat watermelon seeds? Answer: Nothing happens.}). The other input includes the same question, but this time followed by an incorrect answer. We refer to this input as "\textbf{\textit{hallucinated input}}" (e.g., \textit{Question: What happens to you if you eat watermelon seeds? Answer: You grow watermelons in your stomach.}). These two inputs are sent through the LLM individually. When the hallucinated input is processed by the LLM, we get two hidden states, the one relating to the last token of the question part (denoted as $\bm{s_1}$) and the other corresponding to the final token of the entire input (refer to as $\bm{s_2}$). Both $\bm{s_1}$ and $\bm{s_2}$ are derived from the final transformer layer of the language model (e.g., the 32nd layer of LLaMA-2 7B). Likewise, $\bm{s_3}$ is obtained for the correct input. \footnote{Note that there is no need to retrieve the hidden state associated with the question segment again, as it is expected to be identical to $\bm{s_1}$, due to the nature of masked self-attention.} Since $\bm{s_1}$ is extracted after the LLM processing the question but prior to generating a response, it inherently encodes the information regarding an anticipated answer to the given question. In contrast, since $\bm{s_2}$ is acquired after the LLM exposed to a hallucinated response, examining the connection between $\bm{s_1}$ and $\bm{s_2}$ can generate insights into how the model’s final hidden state is influenced by a hallucinated answer. Similarly, tracking the transition from $\bm{s_1}$ to $\bm{s_3}$ can help understand how that is affected by the presence of a correct answer. Therefore, by comparing such pairs of inputs, we can gain insights into how the LLM reacts differently when it encounters a hallucinated response versus a correct one. 

Building upon this experimental framework, we carry out a range of experiments by adjusting the input data, changing the language model, and applying diverse interpretation methods to derive several insightful empirical findings.

\section{Experiments}
\subsection{Experiment Setup}
\label{sec:setup}
\textbf{Datasets.} Our experiments are mainly conducted on two datasets, TruthfulQA \citep{lin2021truthfulqa} and HaluEval \citep{li2023halueval}. Both of them contain QA pairs, where each question is paired with two answers, one being correct and the other incorrect. In particular, TruthfulQA includes a total of 817 samples, which are categorized into two distinct types: "adversarial" (437 samples) and "non-adversarial" (380 samples). Adversarial samples include questions generated through an adversarial procedure, and we experiment with both the entire dataset and each specific type of samples individually. For HaluEval, we use its subset consisting of 10K QA samples generated based on HotpotQA \citep{yang2018hotpotqa}. Each sample contains a question, one correct answer, and one hallucinated answer, but unlike TruthfulQA, each is additionally associated with a text from Wikipedia, which facilitates answering the given question by providing relevant knowledge. In our experiments, we randomly select 1,000 samples from the entire dataset.

\textbf{LLMs.} Our experiments involve using three LLMs, including LLaMA-2 7B, LLaMA-2-Chat 7B, and LLaMA-2 13B \citep{touvron2023llama}. This setup allows us to evaluate and compare the awareness level of hallucination across different model sizes and assess the impact of instruction tuning. The two 7B models comprise 32 layers with a hidden size of 4,096, while the 13B model consists of 40 layers with a hidden size of 5,120. We obtain the models following the instructions provided by Meta AI, \footnote{\url{https://github.com/facebookresearch/llama}} and then implement them using the \texttt{transformers} library. \footnote{\url{https://github.com/huggingface/transformers}}

\begin{figure}[!h]
\centering
\includegraphics[width=0.8\linewidth]{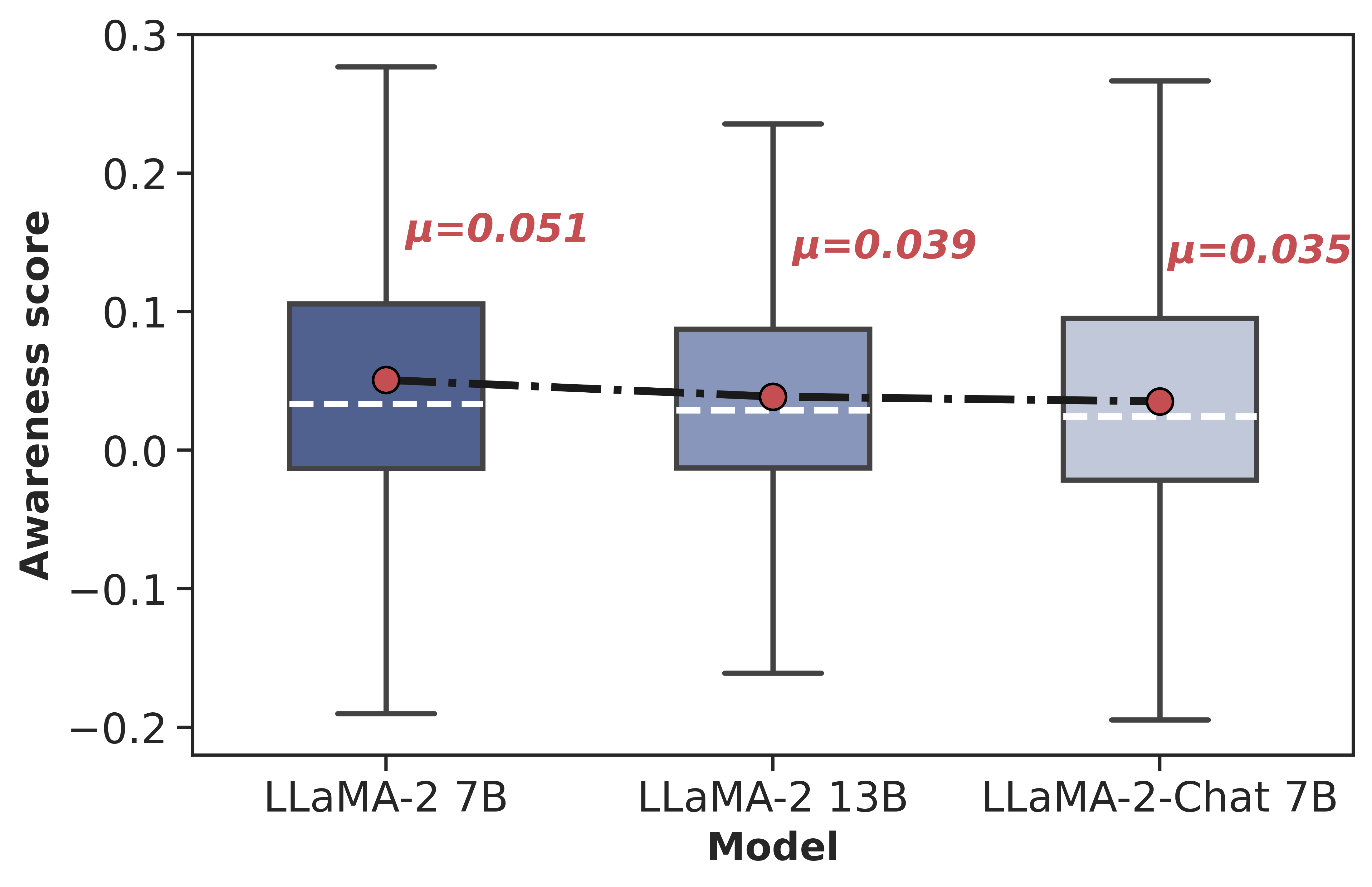}
\caption{Awareness score distributions. We mark the average score for each model in red.}
\label{fig:awareness_models}
\end{figure}

\subsection{Empirical Findings}
\label{sec:findings}
We present our empirical findings as follows.

\textbf{LLM's final hidden state is more susceptible to be influenced by a correct answer compared to a hallucinated one.} Using the TruthfulQA dataset, we conduct experiments following the framework outlined in Figure \ref{fig:illustration}. For each sample, we first calculate two cosine similarity scores: $cos_{halluc} = sim_{cos} (\bm{s_1}, \bm{s_2}) $ and $cos_{corr}=sim_{cos}(\bm{s_1}, \bm{s_3})$. The two scores quantify how much the final hidden state of the LLM is influenced by the hallucinated response and the correct response respectively, conditioned on the same question. We then define \textit{\textbf{awareness score}} as the difference between them: $cos_{halluc}-cos_{corr}$, which quantifies the degree to which the LLM distinguishes hallucinated answers from correct ones. If this score is statistically and significantly non-zero, it indicates that LLMs do possess certain level of hallucination awareness. We present the results in Figure \ref{fig:awareness_models}. First, we consistently observe statistically positive awareness scores across all models, suggesting a noticeable disparity in how the LLM’s final hidden state is influenced by a correct answer compared to a hallucinated one. Particularly, correct response is more likely to change the model’s final hidden state. Second, the level of awareness varies across models, with the highest awareness observed for LLaMA-2 7B and lower levels for its larger and instruction-tuned counterparts. This observation suggests larger and instruction-tuned models are typically more prone to overconfidence, leading them to potentially overlook the generation of hallucinated answers (i.e., less awareness). This aligns with previous research indicating larger and instruction-tuned models often exhibit overconfidence from an uncertainty perspective \citep{duan2023shifting}. Please refer to Appendix \ref{sec:statistical_test_results} for the results of associated statistical tests in Table \ref{tab:awareness_models_test} and \ref{tab:awareness_model_pairs}. 

\begin{figure}[!h]
\centering
\includegraphics[width=0.8\linewidth]{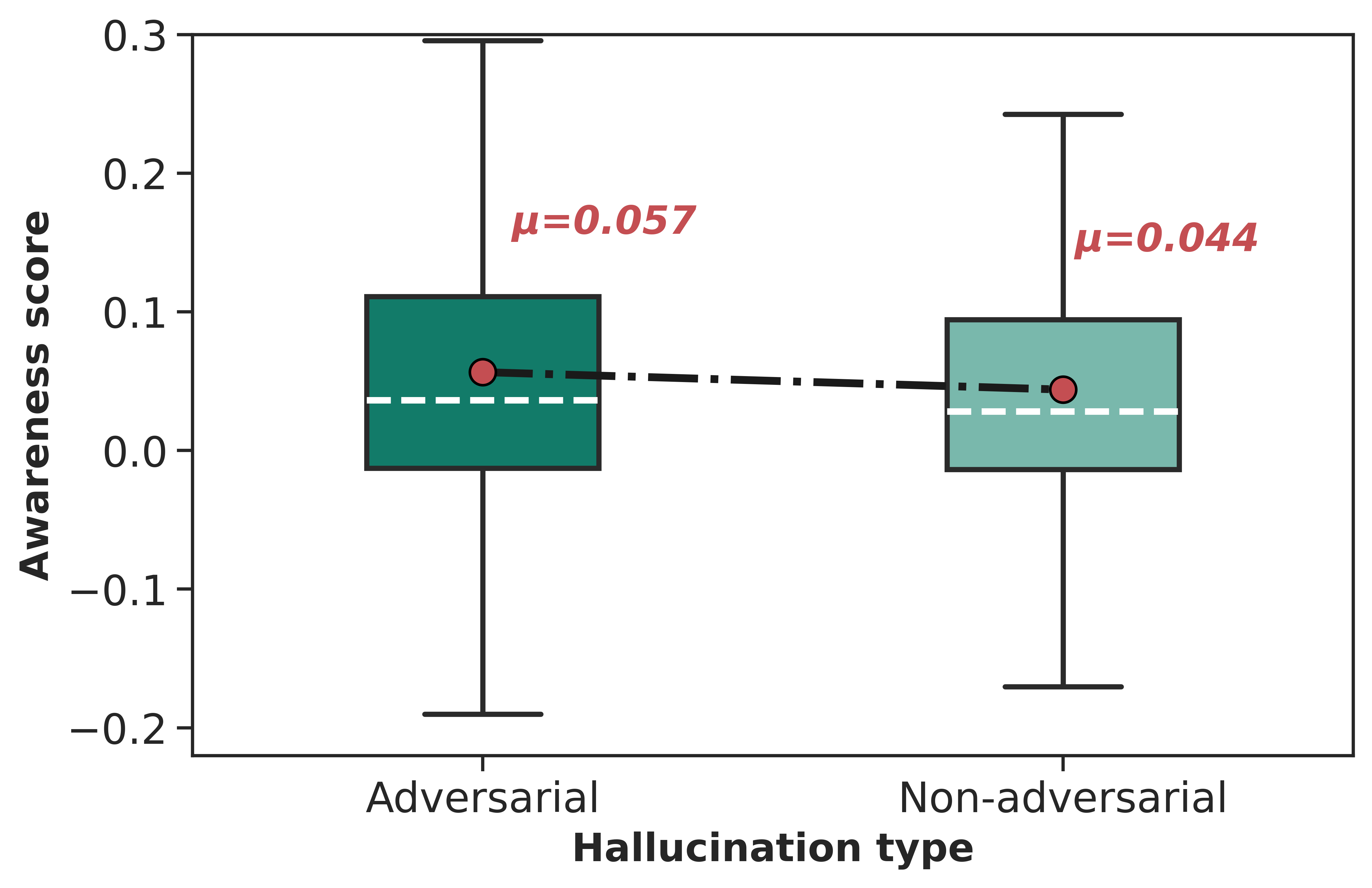}
\caption{Awareness score distributions across hallucination types (LLaMA-2 7B).}
\label{fig:awareness_adversarial}
\end{figure}

\textbf{Strategically inducing an LLM to produce hallucinated responses can raise its awareness consequently.} As the TruthfulQA dataset encompasses both adversarial and non-adversarial samples, we evaluate the LLM’s awareness on each subset following the same procedure. We show the awareness score distributions for LLaMA-2 7B in Figure \ref{fig:awareness_adversarial}, and the corresponding statistical test outcomes are presented in Table \ref{tab:awareness_adversarial} in Appendix \ref{sec:statistical_test_results}. We note the awareness score related to adversarial samples is slightly higher than that of non-adversarial samples, suggesting LLMs tend to exhibit higher level of hallucination awareness when faced with questions that are more likely to be answered incorrectly. We observe similar trends for LLaMA-2-Chat 7B and LLaMA-2 13B, and the corresponding awareness score distributions appear in Figure \ref{fig:awareness_adversarial_chat} and Figure \ref{fig:awareness_adversarial_13B} respectively in Appendix \ref{sec:additional_results}. The associated statistical test results are shown in Table \ref{tab:awareness_adversarial_chat} and \ref{tab:awareness_adversarial_13B} respectively in Appendix \ref{sec:statistical_test_results}.

\textbf{The level of awareness corresponds to the uncertainty of the LLM in the responses.} We define two categories of prompts: \textit{encouraging prompts} and \textit{discouraging prompts}. Encouraging prompts pertain to prompts that reinforce the LLM’s confidence in the responses, such as "\textit{You excel in answering the following question with expertise}". In contrast, discouraging prompts refer to statements like "\textit{You have limited expertise in answering the following question}", which serve to undermine the LLM’s confidence. Given the two categories of prompts, we further define two prompting strategies: \textit{pro-prompting} and \textit{anti-prompting}. Pro-prompting involves placing an encouraging prompt before the question of the correct input, while adding a discouraging prompt preceding the question of the hallucinated input. In contrast, anti-prompting employs an encouraging prompt to prompt the hallucinated input and uses a discouraging prompt to prompt the correct input. Put simply, pro-prompting boosts the LLM's certainty in the accurate answer and introduces skepticism towards the hallucinated response, whereas anti-prompting induces doubt in the correct answer while fostering confidence in the hallucinated one. We examine the LLM’s awareness level under the two prompting strategies and show the results in Figure \ref{fig:awareness_prompting}. Please refer to Table \ref{tab:awareness_prompting} in Appendix \ref{sec:statistical_test_results} for the associated statistical test results. We notice that pro-prompting tends to further raise the LLM’s awareness, whereas anti-prompting may diminish the level of awareness. The observation indicates our derived awareness metric aligns closely with the LLM’s internal confidence in the responses, which is consistent with prior work focused on detecting hallucinations using uncertainty metrics \citep{duan2023shifting}. However, ours offers a new perspective by considering the hidden states. Please refer to Figure \ref{fig:awareness_prompting_chat} (Appendix \ref{sec:additional_results}) and Table \ref{tab:awareness_prompting_chat} (Appendix \ref{sec:statistical_test_results}) for the results of LLaMA-2-Chat 7B.

\begin{figure}[!h]
\centering
\includegraphics[width=0.8\linewidth]{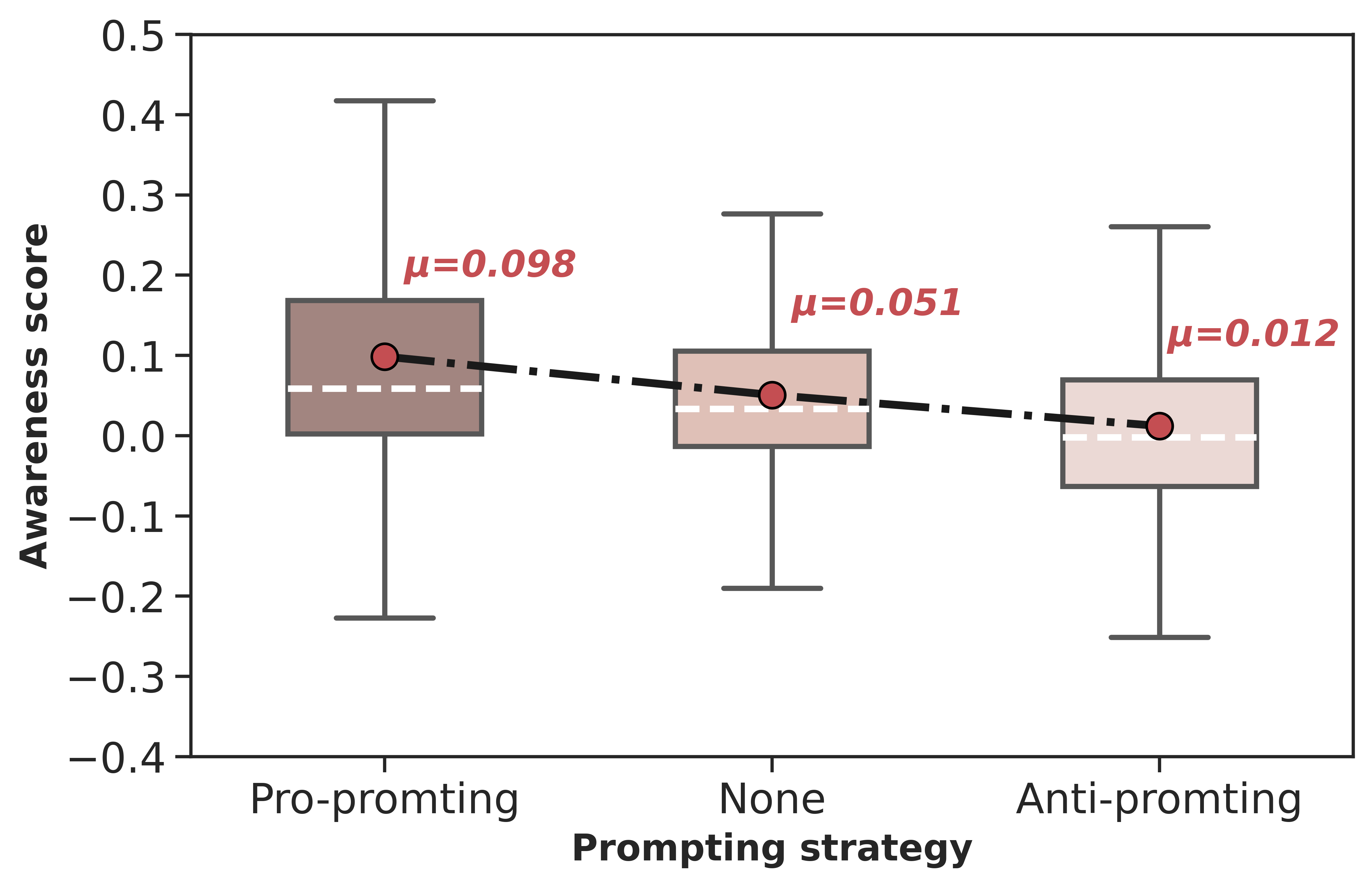}
\caption{Awareness score distributions across prompting strategies (LLaMA-2 7B).}
\label{fig:awareness_prompting}
\end{figure}

\begin{figure}[!h]
\centering
\includegraphics[width=0.8\linewidth]{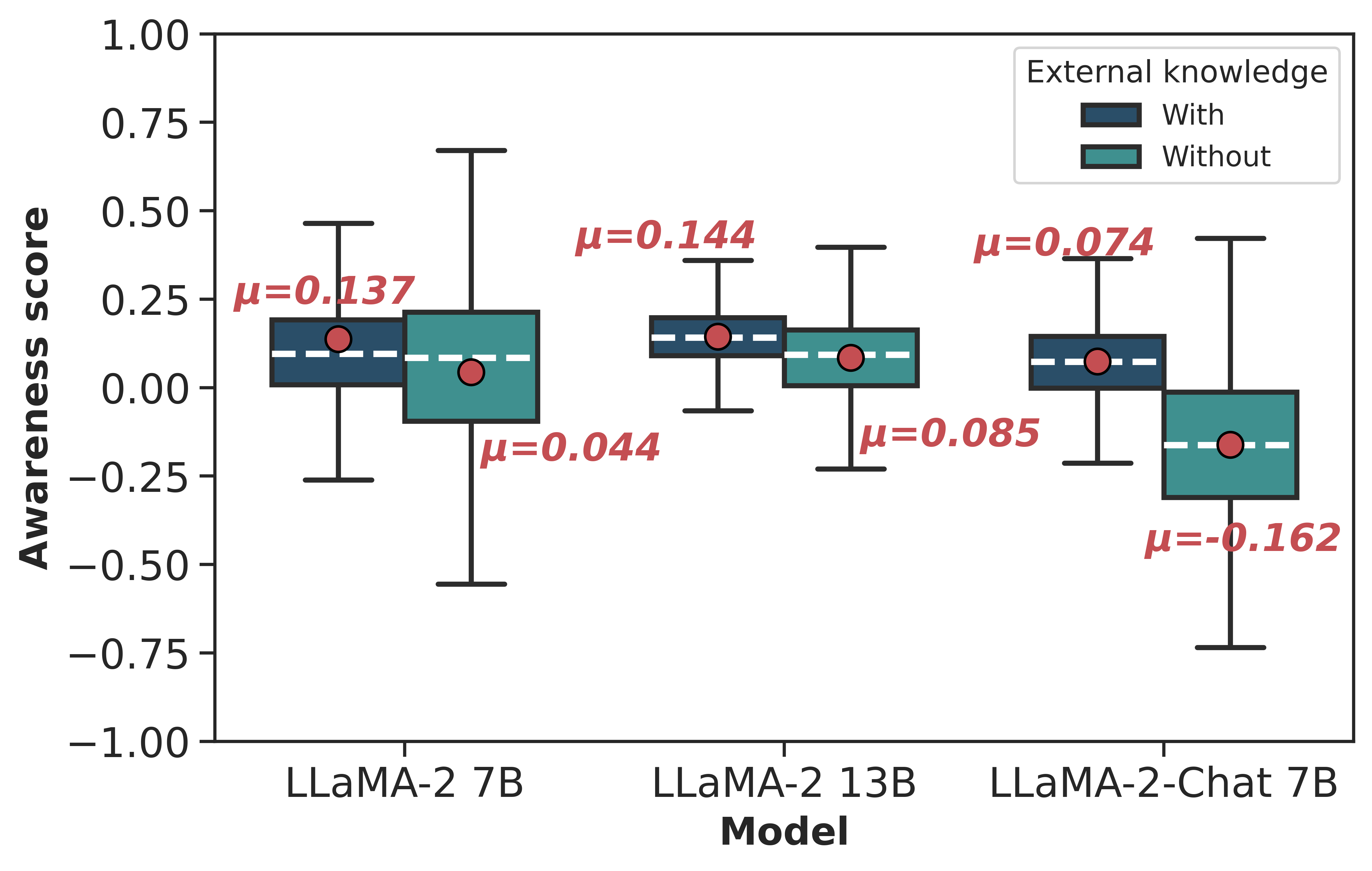}
\caption{Awareness score distributions with and without reference knowledge provided.}
\label{fig:awareness_knowledge}
\end{figure}

\textbf{Providing relevant knowledge can assist LLMs in identifying hallucinated responses.} We conduct experiments and repeat the same procedure to assess the LLM’s awareness of hallucination using the HaluEval dataset (Section \ref{sec:setup}). Unlike the TruthfulQA data, each instance in HaluEval is supplemented with an excerpt from Wikipedia (we refer to it as "external knowledge"), which provides a reference for answering the associated question. This allows us to decide whether or not to include such knowledge into the inputs, and to see if and how the external knowledge influences the LLM’s awareness level. Therefore, for each model, we conduct a pair of experiments, one including the knowledge and the other excluding it. \footnote{When including the knowledge, we directly place the reference text before the question.} We report the awareness score distributions in Figure \ref{fig:awareness_knowledge}. The results show that including reference knowledge can significantly improve the LLM’s awareness of hallucinations, as evidenced by the noticeable discrepancy between each paired boxes, particularly for LLaMA-2-Chat 7B. Please refer to Table \ref{tab:awareness_knowledge} in Appendix \ref{sec:statistical_test_results} for the related statistical tests.

\textbf{The transition between hidden states encodes truthfulness information.} Leveraging the three critical hidden states ($\bm{s_1}$, $\bm{s_2}$, and $\bm{s_3}$) obtained according to Figure \ref{fig:illustration}, we attempt to find two important directions in the hidden representation space, one representing a correct hidden state transition (denoted as "\textit{correct direction}") and the other indicating a hallucinated transition (denoted as "\textit{hallucinated direction}"). We obtain them by three steps. First, for every correct input in the dataset, 
we calculate a vector called "\textit{correct transition vector}" by subtracting $\bm{s_3}$ from $\bm{s_1}$, denoted as $\bm{v_{corr}}=\bm{s_1} - \bm{s_3}$, which tracks the hidden state transition driven by the correct response. Second, we repeat the same procedure for each hallucinated input, yielding the "\textit{hallucinated transition vector}", represented by $\bm{v_{halluc}}=\bm{s_1} - \bm{s_2}$. Third, we employ Principal Component Analysis (PCA) \citep{hotelling1933analysis} technique to identify the first principal component for the correct transition vector ($\bm{v_{corr}}$) and hallucinated transition vector ($\bm{v_{halluc}}$) individually. We name the two principal components as "\textbf{\textit{hallucinated direction}}", denoted as $\bm{d_{halluc}}$ and "\textbf{\textit{correct direction}}", denoted as $\bm{d_{corr}}$. To interpret these two directions, we apply a commonly used vocabulary projection method to project the two directions into the vocabulary space \citep{geva2022transformer, belrose2023eliciting, dar2022analyzing, din2023jump}. Typically, this is approached by computing the dot product between the direction vector and each token embedding (contained in the model’s unembedding matrix). We report the top-10 tokens (ranked by dot product values in descending order) associated with each direction in Table \ref{tab:projection}. Remarkably, we observe that the majority of tokens are related to information truthfulness, with correct-related tokens corresponding to the correct direction and incorrect-relevant tokens associated with the hallucinated direction. This discovery suggests that the transition of final hidden state can encode truthfulness information. We further demonstrate the potential of using the two directions for hallucination mitigation in Section \ref{sec:case_study}.

\begin{table}[!h]
\centering
\renewcommand{\arraystretch}{1}
\scalebox{0.8}{
\begin{tabular}{cc}
\Xhline{1.5pt}
\textbf{Correct direction} & \textbf{Hallucinated direction} \\ \hline
\textit{\textbf{\textcolor{My_color1}{$\_$Correct}}} & \textit{\textbf{\textcolor{My_color2}{$\_$Sad}}} \\
$\_$Register & $\_$Reference \\
\textit{\textbf{\textcolor{My_color1}{$\_$YES}}} & $\_$automatisch \\
\textit{\textbf{\textcolor{My_color1}{$\_$Answer}}} & $\_$Given \\
\textit{\textbf{\textcolor{My_color1}{$\_$Right}}} & $\_$Original \\
$\_$Given & $\_$nearest \\
\textit{\textbf{\textcolor{My_color1}{$\_$answered}}} & $\_$Ev \\
\textit{\textbf{\textcolor{My_color1}{Answer}}} & \textit{\textbf{\textcolor{My_color2}{False}}} \\
$\_$Original & \textit{\textbf{\textcolor{My_color2}{false}}} \\
$\_$Introduction & $\_$Thus \\ 
\Xhline{1.5pt}
\end{tabular}}
\caption{Top-10 tokens associated with each direction. We highlight the tokens related to information truthfulness with \textit{\textbf{\textcolor{My_color1}{green}}} and \textit{\textbf{\textcolor{My_color2}{red}}} for the correct and hallucinated direction respectively. The results are based on LLaMA-2-Chat 7B and the HaluEval dataset.}
\label{tab:projection}
\end{table}

\begin{figure*}[!h]
\captionsetup[subfigure]{justification=centering}
\begin{subfigure}{0.5\textwidth}
   \includegraphics[width=0.8\linewidth]{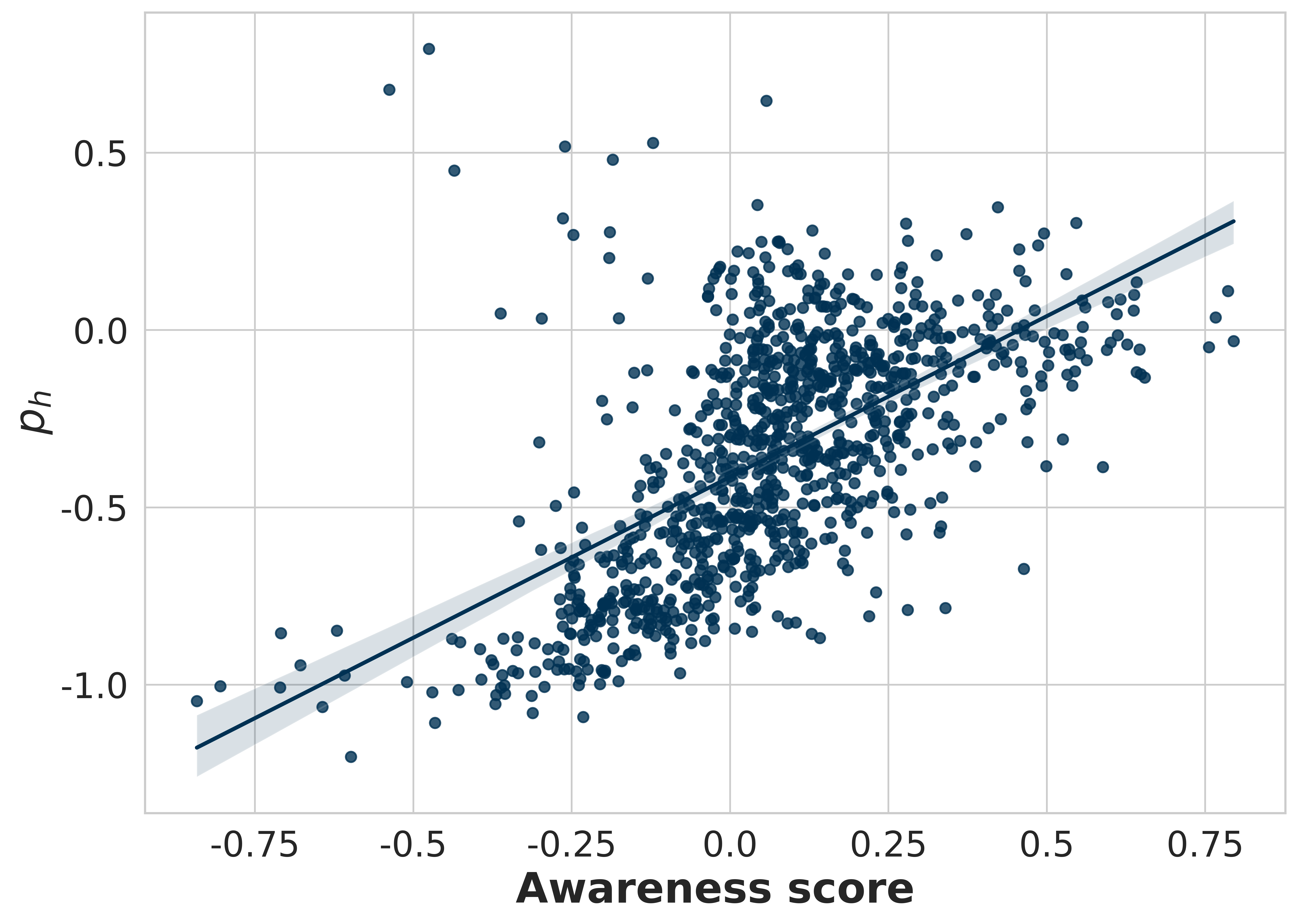}
   \caption{Hallucinated inputs.}
   \label{subfig:regression_hallu}
\end{subfigure}
\begin{subfigure}{0.5\textwidth}
   \includegraphics[width=0.8\linewidth]{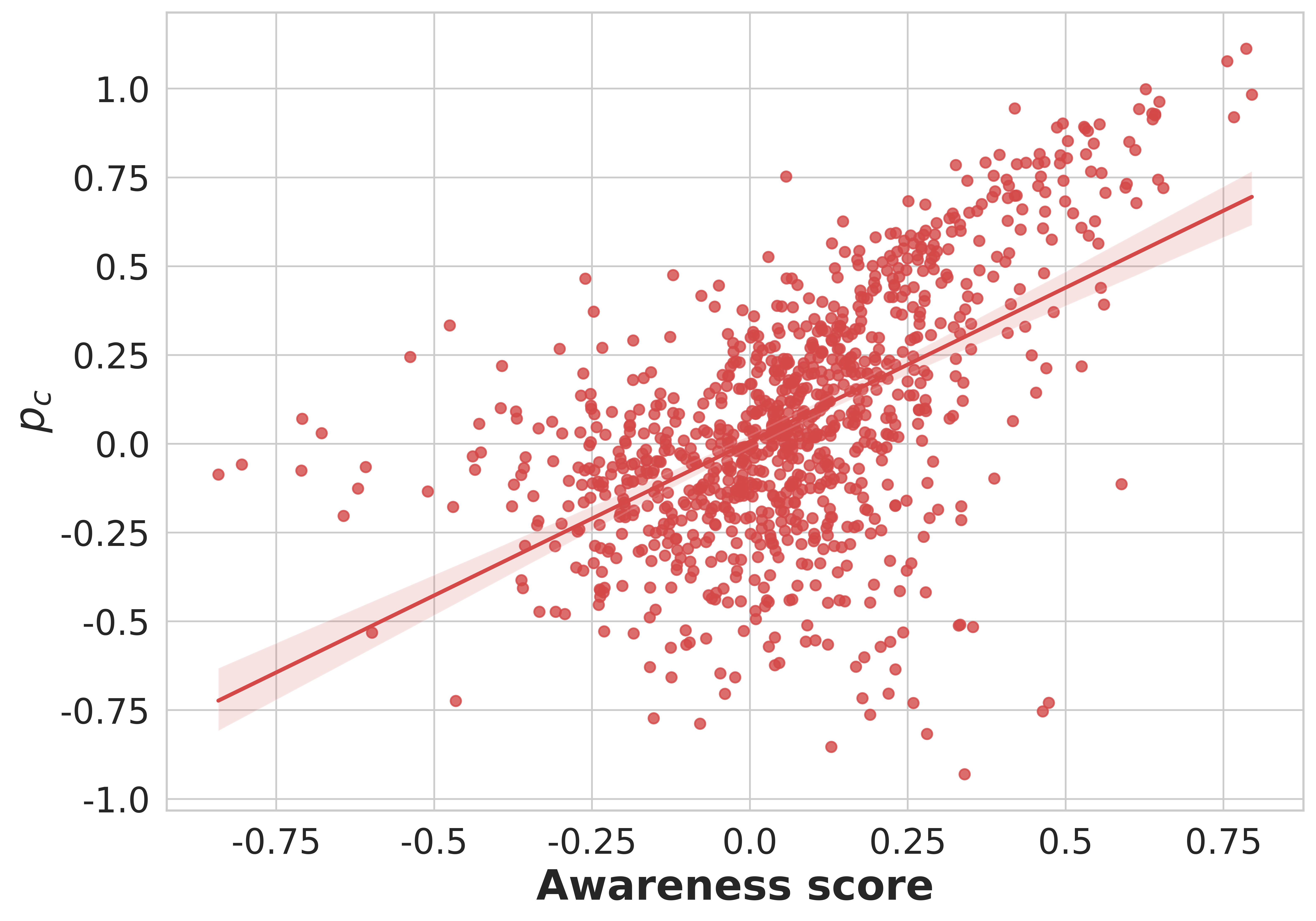}
   \caption{Correct inputs.}
   \label{subfig:regression_corre}
\end{subfigure}
\caption{Regression plots of projection value regressed on the awareness score. The shaded region is the 95\% confidence interval of the regression line. The results are based on the HaluEval data and LLaMA-2-Chat 7B.}
\label{fig:regressions}
\end{figure*}

\textbf{Awareness score aligns closely with the two truthfulness directions.} We explore the relationship between the awareness score and the two directions ($\bm{d_{corr}}$ and $\bm{d_{halluc}}$). We first compute an individual-level awareness for each sample in the dataset (a pair of hallucinated and correct inputs) following the procedure as we describe earlier (Section \ref{sec:findings}). By definition, a higher awareness indicates that distinguishing between this pair of inputs (hallucinated and correct) conditioned on the same question becomes much easier. If so, we anticipate their respective hidden state transition vectors (i.e., $\bm{v_{halluc}}$ and $\bm{v_{corr}}$) should also reflect this ease of differentiation. Specifically, projecting the hallucinated transition vector ($\bm{v_{halluc}}$) onto the hallucinated direction ($\bm{d_{halluc}}$) is expected to yield a larger projection scalar ($p_h$). Similarly, when we project the correct transition vector ($\bm{v_{corr}}$) onto the correct direction ($\bm{d_{corr}}$), we also expect obtaining a larger projection value ($p_c$). We illustrate this idea in Figure \ref{fig:regression_illustration} in Appendix \ref{sec:additional_results}. Formally, to examine the relationship between the awareness score and the projection scalar, we conduct a regression analysis between them. We present the results in Figure \ref{fig:regressions}. Please refer to Table \ref{tab:regression_hallu} and \ref{tab:regression_corre} in Appendix \ref{sec:statistical_test_results} for details. The plots demonstrate a strong \textit{positive correlation} between our awareness metric and the projection scalar, meaning that if a pair of inputs (hallucinated versus correct) are more discernible based on our awareness metric, it suggests their respective transition vectors tend to align more closely with the corresponding direction vectors. It implies the essence captured by the awareness metric can be effectively explained by the two truthfulness directions.

\begin{figure}[!h]
\centering
\includegraphics[width=0.8\linewidth]{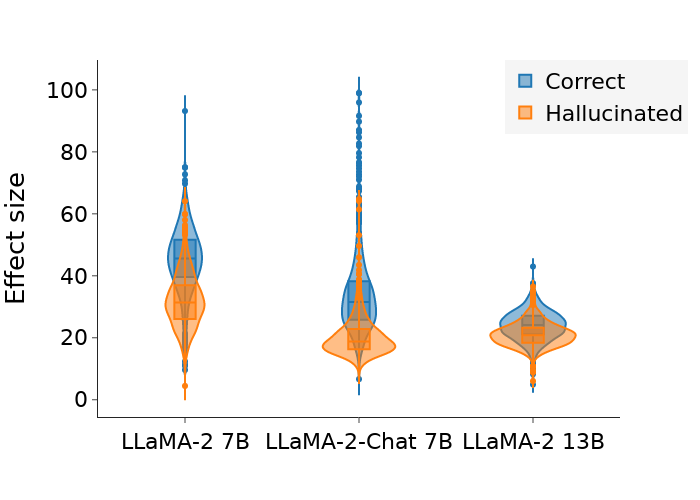}
\caption{Effect size distributions across models. The experiments are conducted on the HaluEval data.}
\label{fig:blocking}
\end{figure}

\textbf{Directly acquiring information from the question component is essential for preventing hallucination.} 
We now take a further step to examine which component \footnote{We consider an input by three components: the reference knowledge component, the question component, and the answer component.} of the input plays a critical role in distinguishing between correct answers and fabricated ones from an information flow perspective \citep{wang2023label}. In this experiment, we particularly focus on the question component. At a higher level, this is done by manipulating the attention edges within the transformer architecture by blocking the attention from the last token to the tokens belonging to the question component. More formally, restricting the attention from position $m$ to position $n$ can be approached by modifying the attention weight matrix $W \in \mathbb{R}^{(d\times d)}$ as: $W_{mn}=-\infty$, where $d$ denotes the length of the input sequence, $m$ and $n$ are the row index and column index of the weight matrix respectively \citep{geva2023dissecting}. \footnote{In our implementation, we set $W_{mn}=-65,504$.} We first consider preventing the last input token from attending to the question tokens in layers above and including the 20$th$ layer. We then define effect size as the degree of change in the final hidden state when attention blocking is implemented ($\bm{s_2^{B}}$ and $\bm{s_3^{B}}$) compared to when it is not ($\bm{s_2}$ and $\bm{s_3}$). More precisely, the effect size is quantified by the L2 norm of the difference vector between the paired hidden states as: $e_{halluc}=\lVert \bm{s_2}-\bm{s_2^{B}}\rVert_2$ and $e_{corr}=\lVert \bm{s_3}-\bm{s_3^{B}}\rVert_2$. Thus, the larger the effect size, the more essential the information directly acquired from the question component becomes, in terms of generating the associated answers. We present the effect size distributions in Figure \ref{fig:blocking} and notice the effect size related to correct inputs (i.e., $e_{corr}$) is considerably and consistently larger than that of hallucinated inputs (i.e., $e_{halluc}$), meaning acquiring information directly from the question is crucial for producing correct answers. In contrast, generating hallucinated responses does not necessitate much question information. This finding implies directly seeking information from the question has potential for mitigating hallucination.

\begin{figure}[!h]
\centering
\includegraphics[width=0.9\linewidth]{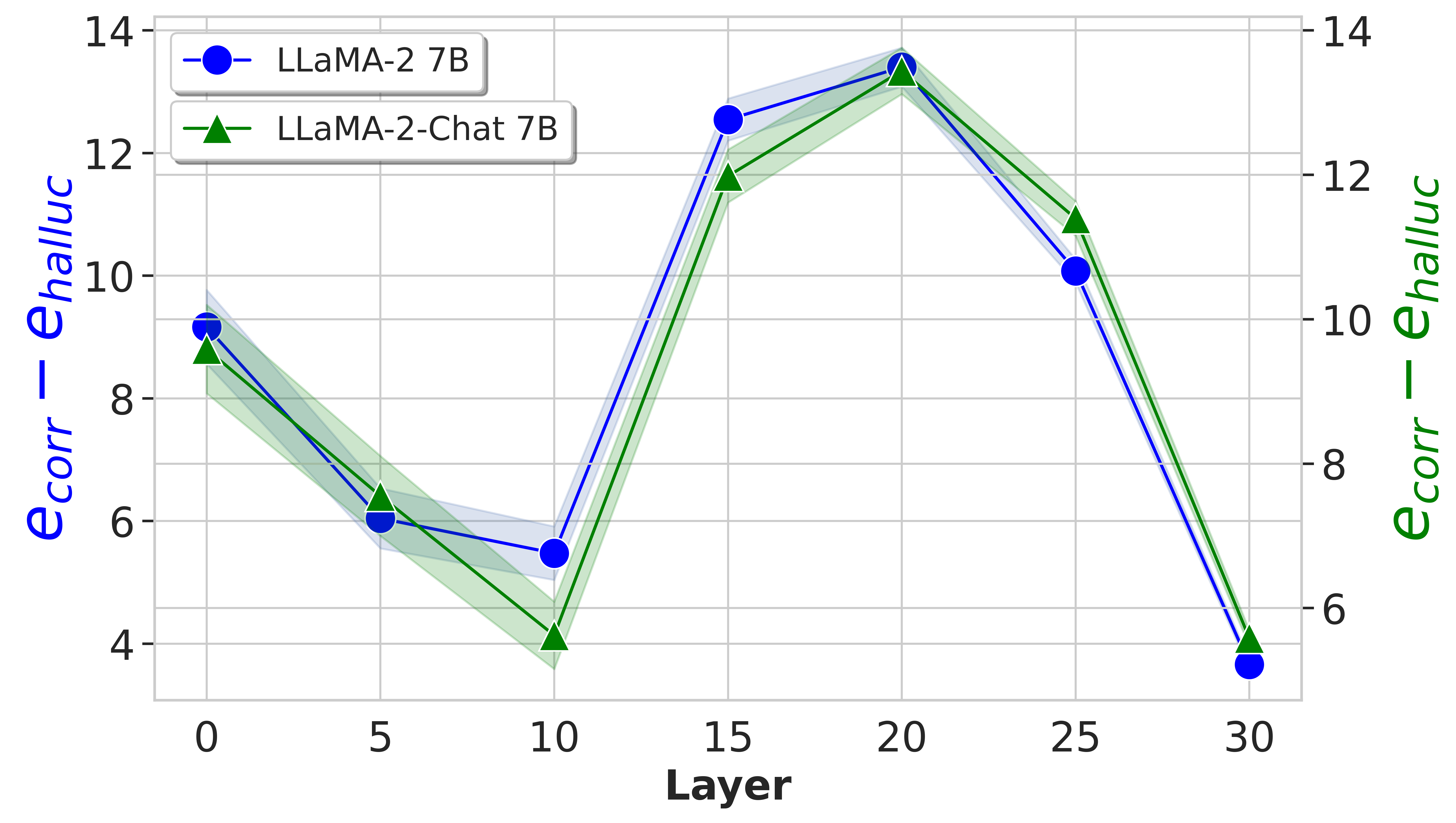}
\caption{Average effect size difference ($e_{corr}-e_{halluc}$) varies across different layers. Each shaded region is the 95\% confidence interval.}
\label{fig:intermediate}
\end{figure}

\textbf{Middle layers in Transformer models are better than other layers at spotting hallucinations.} To examine how the effect size varies across layers, we also explore blocking the attention above (and including) layers 0, 5, 10, 15, 20, 25, and 30 individually. \footnote{The layer index starts from zero.} We present the difference in effect size between the correct and hallucinated inputs (i.e., $e_{corr}-e_{halluc}$) in Figure \ref{fig:intermediate}. The smaller the difference, the more challenging it becomes, to differentiate between hallucinated and correct answers. We observe an $S$-shaped curve, which remains consistent across models. We analyze the curve by three segments: layer 0 to 10, 10 to 20, and 20 to 30. In the first segment, we observe a clear descending trend, indicating that the information contained in layer 0 to 10 does not significantly contribute to distinguishing between hallucinated and correct answers. In the middle segment, the increasing pattern indicates that the information within layers 10 to 20 can help identify hallucinated responses more effectively. In the last segment, starting from layer 20, the difference drops, meaning that the information in the last few layers is less effective in detecting hallucinations compared to middle layers. The findings are consistent with earlier research, which demonstrates that using features extracted from middle layers tends to result in higher accuracies in hallucination detection tasks compared to other layers \citep{azaria2023internal}.

\section{Case Study: Hallucination Mitigation}
\label{sec:case_study}
\begin{table*}[!ht]
\centering
\renewcommand{\arraystretch}{1}
\scalebox{0.58}{
\begin{tabularx}{1.5\textwidth}{lrX}
\Xhline{1.5pt}
$\#1$ &Knowledge: & The 2005 Houston Texans season was the fourth season in franchise history. The team fired head coach Dom Capers after the season; he was replaced by Denver Broncos offensive coordinator Gary Kubiak.Ernest Dominic Capers (born August 7, 1950) is an American football coach and the current defensive coordinator for the Green Bay Packers of the National Football League (NFL). He is the only person to serve two different NFL expansion teams as their inaugural head coach. \\
&Question: & The 2005 Houston Texans fired the only person to serve how many different NFL expansion teams as inaugural head coach? \\
&Original response: & 1 Ernest Dominic Capers \\
&Adjusted response: & 2 \\
&True answer: & two \\ \hline
$\#2$ &Knowledge: & Xiaogan () is a prefecture-level city in east-central Hubei province, People's Republic of China, some 60 km northwest of the provincial capital of Wuhan.Mingguang (), formerly Jiashan County (), is a county-level city of Anhui Province, China. \\
&Question: & Xiaogan and Mingguang, are located in which country? \\
&Original response: & China Xiaogan is a prefecture level city in Hubei province while Mingguang is a county level city in Anhui province \\
&Adjusted response: & China \\
&True answer: & People s Republic of China \\ \hline
$\#3$ &Knowledge: & It has since been broadcast each Christmas season after that as a companion segment in an hour-long slot featuring an unedited version of "A Charlie Brown Christmas". Produced by Lee Mendelson and directed by Bill Melendez, the program made its debut on CBS on December 9, 1965. \\
&Question: & Which one of the Charlie Brown's Christmas Tales was produced by Lee Mendelson? \\
&Original response: & 1965 \\
&Adjusted response: & 1965 s A Charlie Brown Christmas \\
&True answer: & A Charlie Brown Christmas \\
\Xhline{1.5pt}
\end{tabularx}}
\caption{Selected samples where the adjusted response (by adding the offset) better aligns with the ground truth compared to the original response (without the offset). Please refer to Table \ref{tab:alleviation} in Appendix \ref{sec:alleviation} for more examples.}
\label{tab:selected_samples}
\end{table*}

Drawing upon the insights gained from our earlier findings, in this section we conduct a case study to explore the potential to reduce the LLM’s tendency of generating hallucinated answers. In the earlier experiments, we derive two important directions within the LLM’s hidden representation space using PCA: one representing the correct hidden state transition direction (i.e., $\bm{d_{corr}}$) and the other signifying the hallucinated direction (i.e., $\bm{d_{halluc}}$). Here, we attempt to utilize these two identified directions to prevent the LLM from generating hallucinated answers. We draw inspiration from \textit{activation engineering} \citep{subramani2022extracting}, a technique that involves adding vectors to the hidden states of the LLM’s hidden layers to steer its outputs as needed. For example, \citet{konen2024style} explores adding "style vectors" to direct the LLM’s outputs towards a desired style, such as responding positively to a question like "\textit{How is the weather?}" with "\textit{The weather is great!}" rather that "\textit{The weather is bad!}". Moreover, this technique is also employed in language model adaptation by adding logit offsets to guide the LLM’s output as if the model had been fine-tuned \citep{liu2024tuning}. Similarly, here, for every generation step, we propose to add the correct transition direction, $\bm{d_{corr}}$, as an offset to the LLM’s final hidden state \footnote{In the implementation, before adding the vector to the hidden state, we multiply it by a scalar $\alpha=100$, where $\alpha$ serves as a hyperparameter regulating the extent to which the LLM’s output is shifted.}, which is then used to generate the next token. The intuition behind this approach is that the correct hidden state transition direction, $\bm{d_{corr}}$, represents the direction less prone to hallucination; thus, we anticipate that this vector can shift the LLM’s response away from generating hallucinated answers. We present several selected samples where the LLM’s adjusted response (by adding the $\bm{d_{corr}}$ offset) better aligns with the ground truth compared to the original response (without adding the offset) in Table \ref{tab:selected_samples}. When comparing to the original response, we note that adding the $\bm{d_{corr}}$ offset tends to make the adjusted answer align better with the ground truth, exhibiting characteristics such as increased conciseness (as seen in example $\#2$), enhanced completeness ($\#3$), or even a complete reversal from an incorrect answer to a correct one ($\#1$). This case study demonstrates great potential of leveraging guidance extracted from the LLM’s hidden states to mitigate LLM hallucination.

\section{Conclusions}
In this work, we examine the hidden representation space of LLMs and, through carefully designed experiments, we provide empirical evidence suggesting that LLMs do possess awareness of hallucination. We employ several model interpretation techniques to understand how the hidden representations of LLMs react differently to accurate responses compared to hallucinated ones. Informed by the accompanying findings, we show the potential of deriving guidance from the hidden representation space to mitigate LLM hallucination, which could potentially advance the development and adoption of reliable LLMs in critical real-world downstream applications.

\clearpage

\section{Limitations}
This work has several limitations that can be improved in future research. First, we do not differentiate between fine-grained categories of hallucination. Future work can expand upon our experimental framework and delve deeper into understanding how LLMs react to different types of hallucination, such as factual fabrication, instruction inconsistency, logical inconsistency, and so on \citep{rawte2023troubling}. Second, we do not examine the hidden states in the intermediate transformer layers, as only the final hidden state in the last layer directly influences the model’s response. Exploring the intermediate layers in greater depth and investigating the formation of hallucination layer-over-layer may constitute an interesting future direction. Third, our experiments are performed on conventional QA tasks, exploring how to adapt the experimental framework to examine hallucination in more complicated or domain specific tasks - such as numerical reasoning or financial text comprehension - and potentially extending the analysis to include multimodal features, warrants further investigation. Finally, given that our study generates several empirical findings regarding how LLMs perceive hallucinated responses, future research may be needed to develop more effective strategies for mitigating LLM hallucination, drawing insights from the observations made in this study.

\bibliography{custom}

\appendix
\section{Results of Statistical Tests}
\label{sec:statistical_test_results}
To assess the statistical significance of the awareness score being greater than zero, we conduct pairwise one-tailed t-test for each model. Before performing each t-test, we ensure that the assumptions required for using the t-test, such as the normality, are met properly. We present the results in Table \ref{tab:awareness_models_test}. Additionally, we compare the difference in awareness scores between each pair of models and present the results in Table \ref{tab:awareness_model_pairs}.

We assess the statistical significance of the awareness score being above zero for adversarial and non-adversarial samples individually, and present the results for LLaMA-2 7B in Table \ref{tab:awareness_adversarial}. Please refer to Table \ref{tab:awareness_adversarial_chat} and Table \ref{tab:awareness_adversarial_13B} for the results of LLaMA-2-Chat 7B and LLaMA-2 13B respectively.

The statistical test results for exploring different prompting strategies are reported in Table \ref{tab:awareness_prompting} and \ref{tab:awareness_prompting_chat}.

The detailed regression results (projection value regressed on awareness score) are presented in Table \ref{tab:regression_hallu} and \ref{tab:regression_corre}.

\begin{table}[!h]
\centering
\renewcommand{\arraystretch}{1}
\scalebox{0.6}{
\begin{tabular}{rccc}
\Xhline{1.5pt}
 & \textbf{LLaMA-2 7B} & \textbf{LLaMA-2 13B} & \textbf{LLaMA-2-Chat 7B} \\ \hline
Awareness score & 0.051*** & 0.039*** & 0.035*** \\
t-statistic & 10.69 & 12.27 & 8.47 \\
p-value & 2.42E-25 & 3.37E-32 & 5.81E-17 \\
df & 816 & 816 & 816 \\
\Xhline{1.5pt}
\end{tabular}}
\caption{Results of pairwise one-tailed t-tests. The null hypothesis is the awareness score is less than or equal to zero. $^{***}p<0.01$; $^{**}p<0.05$; $^{*}p<0.1$.}
\label{tab:awareness_models_test}
\end{table}

\begin{table*}[!ht]
\centering
\renewcommand{\arraystretch}{1}
\scalebox{0.7}{
\begin{tabular}{r|cc|cc|cc}
\Xhline{1.5pt}
Model & \multicolumn{2}{c|}{\textbf{LLaMA-2 7B}} & \multicolumn{2}{c|}{\textbf{LLaMA-2 13B}} & \multicolumn{2}{c}{\textbf{LLaMA-2-Chat 7B}} \\ \hline
External knowledge & w/ & w/o & w/ & w/o & w/ & w/o \\
Awareness score & 0.137*** & 0.044*** & 0.144*** & 0.085*** & 0.074*** & -0.162 (ns) \\
t-statistic & 18.04 & 5.02 & 47.03 & 19.13 & 16.71 & -23.94 \\
p-value & 1.62E-63 & 3.09E-07 & 7.74E-256 & 4.93E-70 & 9.77E-56 & 1 \\
df & 999 & 999 & 999 & 999 & 999 & 999 \\ 
\Xhline{1.5pt}
\end{tabular}}
\caption{Results of pairwise one-tailed t-tests. The null hypothesis is the awareness score is less than or equal to zero. $^{***}p<0.01$; $^{**}p<0.05$; $^{*}p<0.1$.}
\label{tab:awareness_knowledge}
\end{table*}

\begin{table}[!h]
\centering
\renewcommand{\arraystretch}{1}
\scalebox{0.65}{
\begin{tabular}{rcc}
\Xhline{1.5pt}
 &$\bm{\mu}$ \textbf{(7B)} - $\bm{\mu}$ \textbf{(13B)} & $\bm{\mu}$ \textbf{(7B)} - $\bm{\mu}$ \textbf{(7B-Chat)} \\ \hline
Score difference & 0.012** & 0.015*** \\
t-statistic & 3.01 & 3.41 \\
p-value & 1.36E-03 & 3.39E-04 \\
df & 816 & 816 \\ 
\Xhline{1.5pt}
\end{tabular}}
\caption{Results of pairwise one-tailed t-tests. The null hypothesis is the awareness score difference is less than or equal to zero. $^{***}p<0.01$; $^{**}p<0.05$; $^{*}p<0.1$.}
\label{tab:awareness_model_pairs}
\end{table}

\begin{table}[!h]
\centering
\renewcommand{\arraystretch}{1}
\scalebox{0.65}{
\begin{tabular}{rcc}
\Xhline{1.5pt}
 & \textbf{Adversarial} & \textbf{Non-adversarial} \\ \hline
Awareness score & 0.057*** & 0.044*** \\
t-statistic & 8.07 & 7.04 \\
p-value & 3.47E-15 & 4.44E-12 \\
df & 436 & 379 \\ 
\Xhline{1.5pt}
\end{tabular}}
\caption{Results of pairwise one-tailed t-tests (LLaMA-2 7B). The null hypothesis is the awareness score is less than or equal to zero. $^{***}p<0.01$; $^{**}p<0.05$; $^{*}p<0.1$.}
\label{tab:awareness_adversarial}
\end{table}

\begin{table}[!h]
\centering
\renewcommand{\arraystretch}{1}
\scalebox{0.65}{
\begin{tabular}{rcc}
\Xhline{1.5pt}
& \textbf{Adversarial} & \textbf{Non-adversarial} \\ \hline
Awareness score & 0.041*** & 0.029*** \\
t-statistic & 6.84 & 5.03 \\
p-value & 1.35E-11 & 3.79E-07 \\
df & 436 & 379 \\ 
\Xhline{1.5pt}
\end{tabular}}
\caption{Results of pairwise one-tailed t-tests (LLaMA-2-Chat 7B). The null hypothesis is the awareness score is less than or equal to zero. $^{***}p<0.01$; $^{**}p<0.05$; $^{*}p<0.1$.}
\label{tab:awareness_adversarial_chat}
\end{table}

\begin{table}[!h]
\centering
\renewcommand{\arraystretch}{1}
\scalebox{0.65}{
\begin{tabular}{rcc}
\Xhline{1.5pt}
& \textbf{Adversarial} & \textbf{Non-adversarial} \\ \hline
Awareness score & 0.044*** & 0.032*** \\
t-statistic & 9.64 & 7.63 \\
p-value & 2.32E-20 & 9.52E-14 \\
df & 436 & 379 \\
\Xhline{1.5pt}
\end{tabular}}
\caption{Results of pairwise one-tailed t-tests (LLaMA-2 13B). The null hypothesis is the awareness score is less than or equal to zero. $^{***}p<0.01$; $^{**}p<0.05$; $^{*}p<0.1$.}
\label{tab:awareness_adversarial_13B}
\end{table}

\begin{table}[!h]
\centering
\renewcommand{\arraystretch}{1}
\scalebox{0.65}{
\begin{tabular}{rccc}
\Xhline{1.5pt}
&\textbf{Pro-prompting} & \textbf{None} & \textbf{Anti-prompting} \\ \hline
Awareness score & 0.098*** & 0.051*** & 0.012* \\
t-statistic & 17.47 & 10.69 & 2.20 \\
p-value & 1.25E-58 & 2.42E-25 & 1.40E-02 \\
df & 816 & 816 & 816 \\
\Xhline{1.5pt}
\end{tabular}}
\caption{Results of pairwise one-tailed t-tests across prompting strategies (LLaMA-2 7B). The null hypothesis is the awareness score is less than or equal to zero. $^{***}p<0.01$; $^{**}p<0.05$; $^{*}p<0.1$.}
\label{tab:awareness_prompting}
\end{table}

\begin{table}[!h]
\centering
\renewcommand{\arraystretch}{1}
\scalebox{0.65}{
\begin{tabular}{rccc}
\Xhline{1.5pt}
& \textbf{Pro-prompting} & \textbf{None} & \textbf{Anti-prompting} \\ \hline
Awareness score & 0.087*** & 0.035*** & -0.014 (ns) \\
t-statistic & 19.80 & 8.47 & -3.08 \\
p-value & 7.62E-72 & 5.81E-17 & 9.99E-01 \\
df & 816 & 816 & 816 \\
\Xhline{1.5pt}
\end{tabular}}
\caption{Results of pairwise one-tailed t-tests across prompting strategies (LLaMA-2-Chat 7B). The null hypothesis is the awareness score is less than or equal to zero. $^{***}p<0.01$; $^{**}p<0.05$; $^{*}p<0.1$.}
\label{tab:awareness_prompting_chat}
\end{table}

\begin{table}[!h] \centering
\scalebox{0.7}{
\begin{tabular}{@{\extracolsep{5pt}}lc}
\\[-1.8ex]\hline
\hline \\[-1.8ex]
& \multicolumn{1}{c}{\textit{Dependent variable: $p_{h}$}} \
\cr \cline{2-2}
\\[-1.8ex] & (1) \\
\hline \\[-1.8ex]
 Awareness score & 0.907$^{***}$ \\
& (0.038) \\
 const & -0.414$^{***}$ \\
& (0.009) \\
\hline \\[-1.8ex]
 Observations & 1000 \\
 $R^2$ & 0.362 \\
 Adjusted $R^2$ & 0.362 \\
 Residual Std. Error & 0.270 (df=998) \\
 F Statistic & 567.406$^{***}$ (df=1; 998) \\
\hline
\hline \\[-1.8ex]
\textit{Note:} & \multicolumn{1}{r}{$^{*}$p$<$0.1; $^{**}$p$<$0.05; $^{***}$p$<$0.01} \\
\end{tabular}}
\caption{Regression results for hallucinated inputs (associated with Figure \ref{subfig:regression_hallu}).}
\label{tab:regression_hallu}
\end{table}

\begin{table}[!h] \centering
\scalebox{0.7}{
\begin{tabular}{@{\extracolsep{5pt}}lc}
\\[-1.8ex]\hline
\hline \\[-1.8ex]
& \multicolumn{1}{c}{\textit{Dependent variable: $p_{c}$}} \
\cr \cline{2-2}
\\[-1.8ex] & (1) \\
\hline \\[-1.8ex]
 Awareness score & 0.867$^{***}$ \\
& (0.040) \\
 const & 0.006$^{}$ \\
& (0.009) \\
\hline \\[-1.8ex]
 Observations & 1000 \\
 $R^2$ & 0.323 \\
 Adjusted $R^2$ & 0.322 \\
 Residual Std. Error & 0.281 (df=998) \\
 F Statistic & 476.223$^{***}$ (df=1; 998) \\
\hline
\hline \\[-1.8ex]
\textit{Note:} & \multicolumn{1}{r}{$^{*}$p$<$0.1; $^{**}$p$<$0.05; $^{***}$p$<$0.01} \\
\end{tabular}}
\caption{Regression results for correct inputs (associated with Figure \ref{subfig:regression_corre}).}
\label{tab:regression_corre}
\end{table}

\clearpage
\section{Additional Results}
\label{sec:additional_results}
\begin{figure}[!h]
\centering
\includegraphics[width=0.76\linewidth]{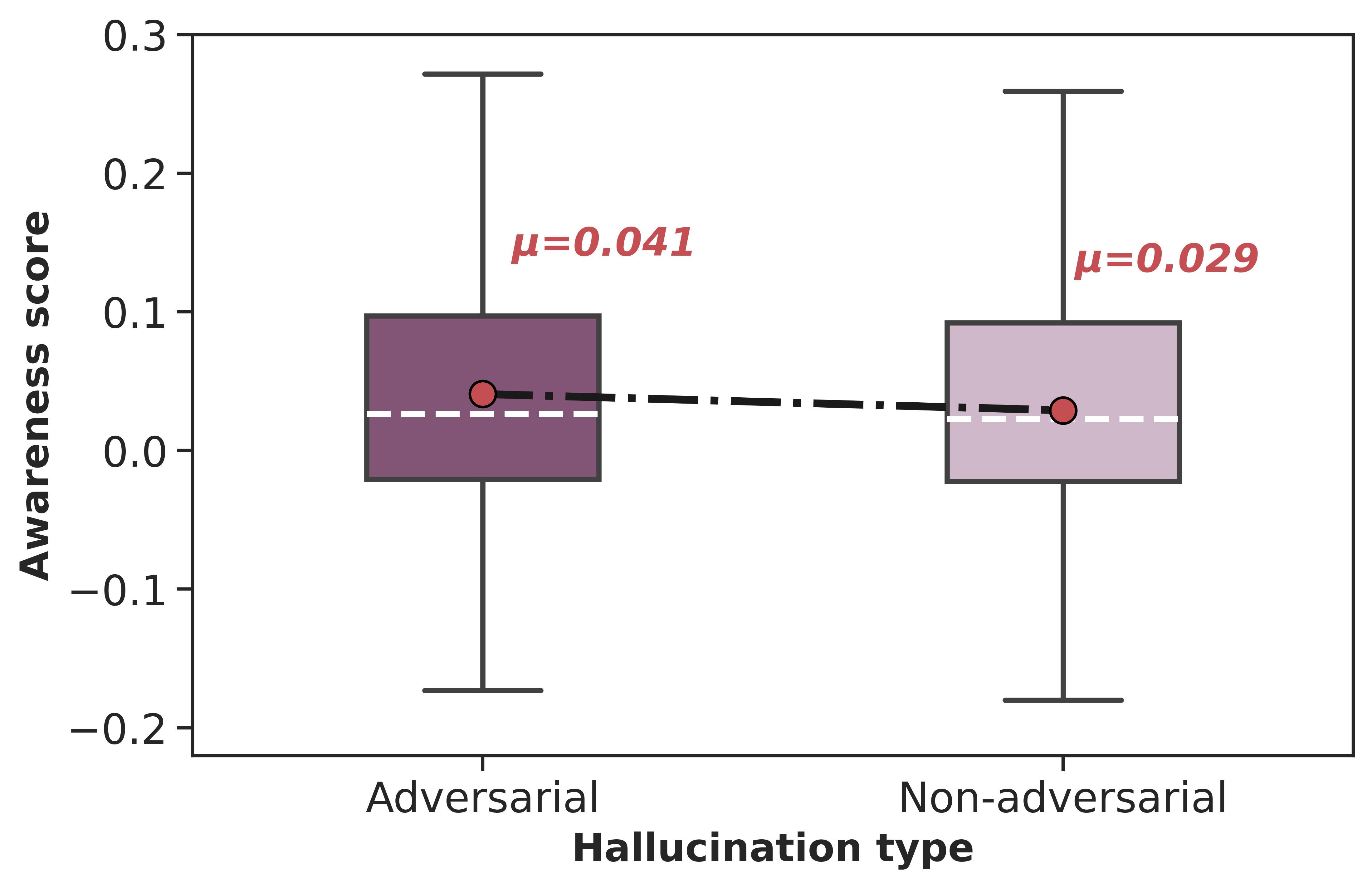}
\caption{Awareness score distribution across hallucination types (LLaMA-2-Chat 7B).}
\label{fig:awareness_adversarial_chat}
\end{figure}

\begin{figure}[!h]
\centering
\includegraphics[width=0.76\linewidth]{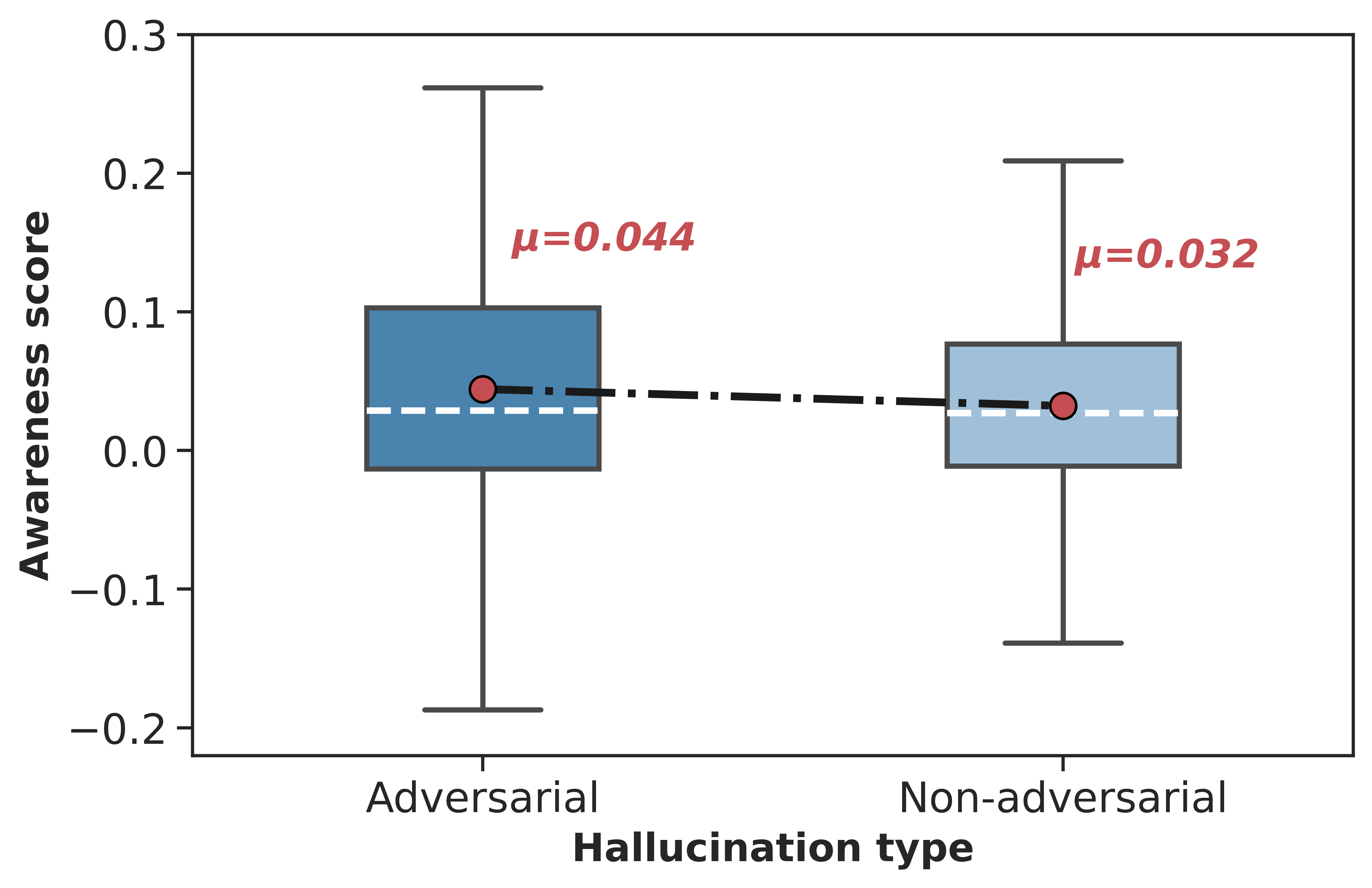}
\caption{Awareness score distribution across hallucination types (LLaMA-2 13B).}
\label{fig:awareness_adversarial_13B}
\end{figure}

\begin{figure}[!h]
\centering
\includegraphics[width=0.76\linewidth]{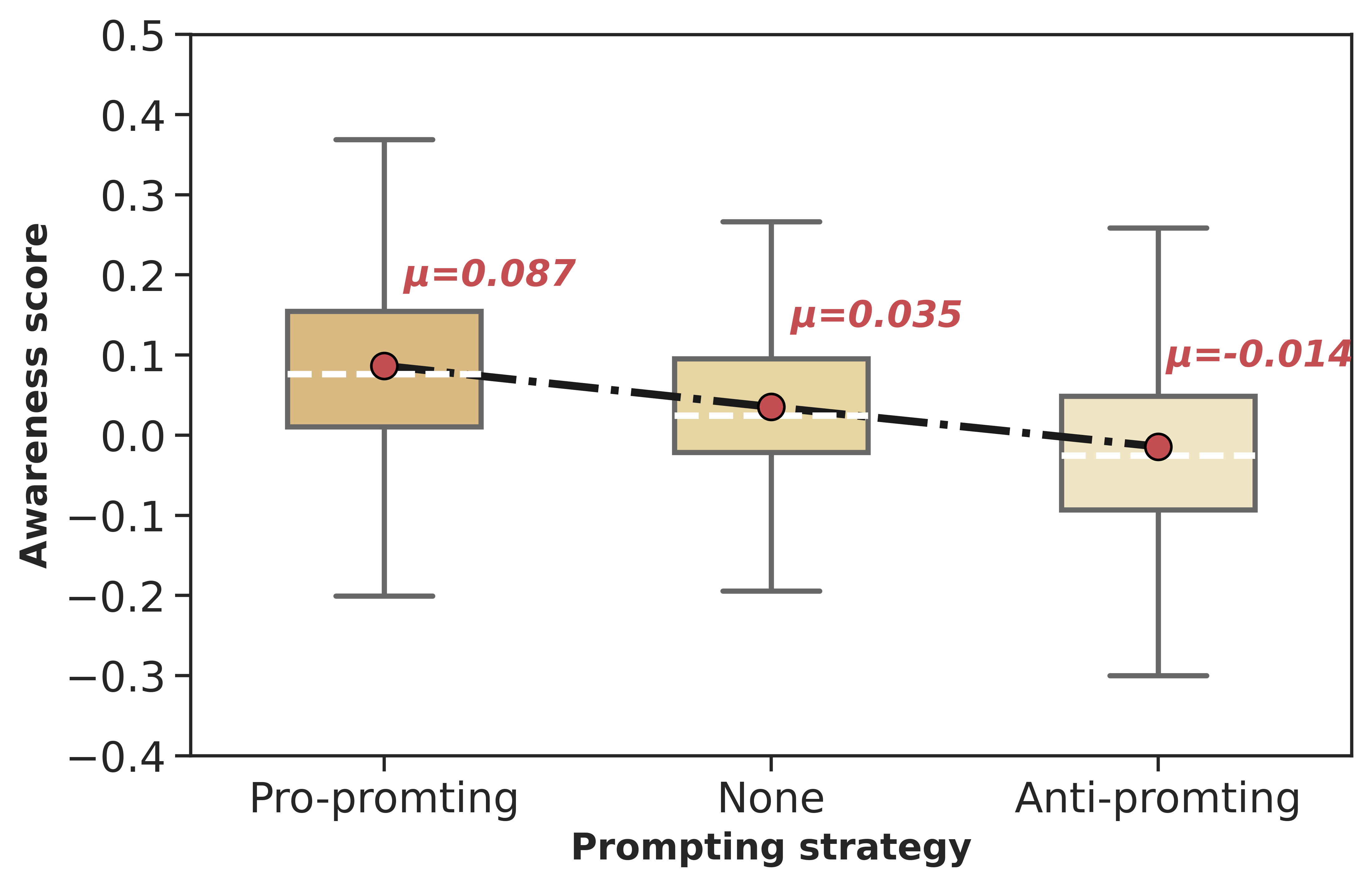}
\caption{Awareness score distributions across prompting strategies (LLaMA-2-Chat 7B).}
\label{fig:awareness_prompting_chat}
\end{figure}

\begin{figure}[!h]
\centering
\includegraphics[width=0.6\linewidth]{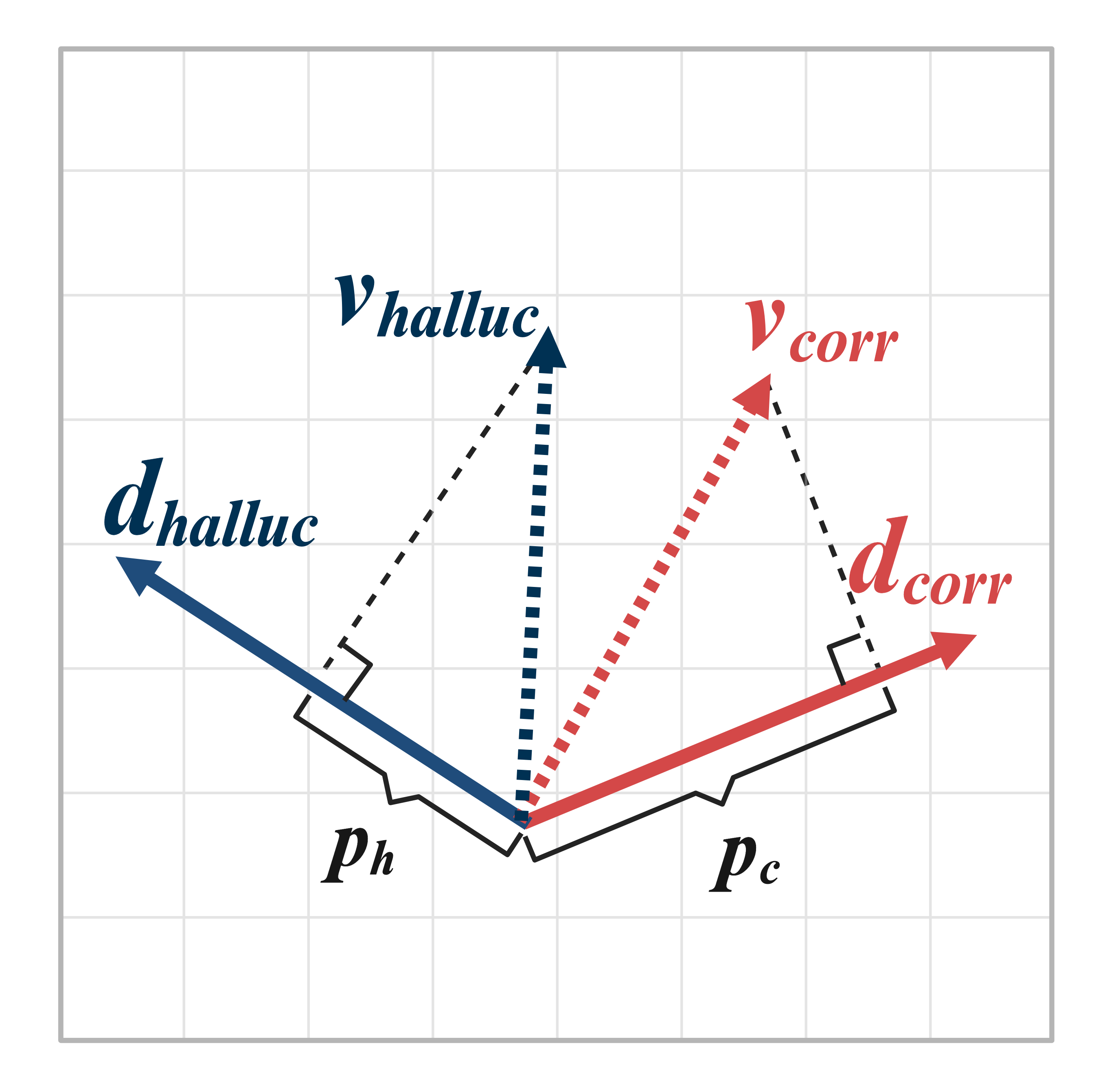}
\caption{Projection illustration. We project the hidden state transition vectors ($\bm{v_{halluc}}$ and $\bm{v_{corr}}$) onto their respective direction vectors ($\bm{d_{halluc}}$ and $\bm{d_{corr}}$), and obtain two scalar projections ($p_h$ and $p_c$).}
\label{fig:regression_illustration}
\end{figure}

The awareness score distributions across hallucination types for LLaMA-2-Chat 7B and LLaMA-2 13B are shown in Figure \ref{fig:awareness_adversarial_chat} and Figure \ref{fig:awareness_adversarial_13B}, and the results associated with prompting strategies for LLaMA-2-Chat 7B appear in Figure \ref{fig:awareness_prompting_chat}. The projection illustration is shown in Figure \ref{fig:regression_illustration}.

\clearpage
\section{Hallucination Mitigation}
\label{sec:alleviation}
\begin{table}[!ht]
\centering
\renewcommand{\arraystretch}{1}
\scalebox{0.6}{
\begin{tabularx}{1.5\textwidth}{lrX}
\Xhline{1.5pt}
$\#1$ &Knowledge: & Gigantic is a 2008 independent comedy film directed by Matt Aselton and starring Paul Dano, Zooey Deschanel, John Goodman, Edward Asner and Jane Alexander. Early in his career, he was best known for playing Dan Conner on the ABC TV series "Roseanne" (1988–1997), for which he won a Golden Globe Award for Best Actor in 1993. \\
&Question: & The Golden Globe Award winner for best actor from "Roseanne" starred along what actress in Gigantic? \\
&Original response: & John Goodman \\
&Adjusted response: & John Goodman starred along Zooey Deschanel in Gigantic \\
&True answer: & Zooey Deschanel \\ \hline
$\#2$ &Knowledge: & John Randal Kleiser (born July 20, 1946) is an American film director and producer, best known for directing the 1978 musical romantic comedy film "Grease".Kyle Schickner is an American film producer, writer, director, actor and a bisexual civil rights activist. \\
&Question: & Are Randal Kleiser and Kyle Schickner of the same nationality? \\
&Original response: & No Randal Kleiser is American and Kyle Schickner is American \\
&Adjusted response: & Yes Randal Kleiser and Kyle Schickner are both American \\
&True answer: & yes \\ \hline
$\#3$ &Knowledge: & Symbolic Manipulation Program, usually called SMP, was a computer algebra system designed by Chris A. Cole and Stephen Wolfram at Caltech circa 1979 and initially developed in the Caltech physics department under Wolfram's leadership with contributions from Geoffrey C. Fox, Jeffrey M. Greif, Eric D. Mjolsness, Larry J. Romans, Timothy Shaw, and Anthony E. Terrano. He is the author of the book "A New Kind of Science." \\
&Question: & The author of the book "A New Kind of Science" designed a computer algebra system at which university? \\
&Original response: & 1 Caltech California Institute of Technology 2 MIT Massachusetts Institute of Technology 3 Harvard University 4 Stanford University \\
&Adjusted response: & 1979 Caltech \\
&True answer: & Caltech \\ \hline
$\#4$ &Knowledge: & Xiaogan () is a prefecture-level city in east-central Hubei province, People's Republic of China, some 60 km northwest of the provincial capital of Wuhan.Mingguang (), formerly Jiashan County (), is a county-level city of Anhui Province, China. \\
&Question: & Xiaogan and Mingguang, are located in which country? \\
&Original response: & China Xiaogan is a prefecture level city in Hubei province while Mingguang is a county level city in Anhui province \\
&Adjusted response: & China \\
&True answer: & People s Republic of China \\ \hline
$\#5$ &Knowledge: & Miss Perú 2010, the 58th Miss Perú pageant was held at the on May 22, 2010.Miss Universe 2010, the 59th Miss Universe pageant, was held on 23 August 2010 at Mandalay Bay Events Center in Las Vegas, Nevada, USA. \\
&Question: & Which was around longer in 2010, Miss Universe or Miss Perú? \\
&Original response: & Miss Perú \\
&Adjusted response: & Miss Universe \\
&True answer: & Miss Universe \\ \hline
$\#6$ &Knowledge: & Nemesis is the first rap crew from Dallas, Texas to enjoy popularity in the US and world-wide, thanks to their label, Profile Records, which was well known for bringing out many popular east coast rap releases such as Run-DMC, Dr. Jeckyll \& Mr. Hyde, and Dana Dane.Run-D.M.C. was an American hip hop group from Hollis, Queens, New York, founded in 1981 by Joseph Simmons, Darryl McDaniels, and Jason Mizell. Run-D.M.C. Run-D.M.C. Run-D.M.C. Run-D.M.C. \\
&Question: & What hip hop group from Hollis, Queens, New York, founded in 1981 by Joseph Simmons, Darryl McDaniels, and Jason Mizell, was brought out by  the first rap crew from Dallas, Texas? \\
&Original response: & Nemesis \\
&Adjusted response: & Run D M C \\
&True answer: & Run D M C \\ \hline
$\#7$ &Knowledge: & Heaven Is a Playground is a 1991 film based on Rick Telander's book of the same name. It was written and directed by Randall Fried, and stars Michael Warren, D.B. Sweeney, Victor Love, Bo Kimble, Richard Jordan and Janet Julian. Telander himself has a cameo appearance in the movie.Richard Jordan (born Robert Anson Jordan, Jr.; July 19, 1937 – August 30, 1993) was an American stage, screen, and television actor. A long-time member of the New York Shakespeare Festival, he performed in many Off Broadway and Broadway plays. \\
&Question: & Who has a cameo appearance in the film in which both Bo Kimble and a member of the New York Shakespeare Festival who has performed in many Broadway plays star in? \\
&Original response: & Randall Fried \\
&Adjusted response: & Rick Telander \\
&True answer: & Rick Telander \\ \hline
$\#8$ &Knowledge: & The 2005 Houston Texans season was the fourth season in franchise history. The team fired head coach Dom Capers after the season; he was replaced by Denver Broncos offensive coordinator Gary Kubiak.Ernest Dominic Capers (born August 7, 1950) is an American football coach and the current defensive coordinator for the Green Bay Packers of the National Football League (NFL). He is the only person to serve two different NFL expansion teams as their inaugural head coach. \\
&Question: & The 2005 Houston Texans fired the only person to serve how many different NFL expansion teams as inaugural head coach? \\
&Original response: & 1 Ernest Dominic Capers \\
&Adjusted response: & 2 \\
&True answer: & two \\ \hline
$\#9$ &Knowledge: & Yellowcard was an American pop punk band that formed in Jacksonville, Florida, in 1997 and were based in Los Angeles beginning in 2000.For Against is a United States post-punk/dream pop band from Lincoln, Nebraska. \\
&Question: & Were Yellowcard and For Against both American pop bands? \\
&Original response: & Yes Both Yellowcard and For Against are American pop bands Yellowcard was formed in Jacksonville Florida and later based in Los Angeles while For Against was formed in Lincoln Nebraska \\
&Adjusted response: & Yes \\
&True answer: & yes \\ \hline
$\#10$ &Knowledge: & It has since been broadcast each Christmas season after that as a companion segment in an hour-long slot featuring an unedited version of "A Charlie Brown Christmas". Produced by Lee Mendelson and directed by Bill Melendez, the program made its debut on CBS on December 9, 1965. \\
&Question: & Which one of the Charlie Brown's Christmas Tales was produced by Lee Mendelson? \\
&Original response: & 1965 \\
&Adjusted response: & 1965 s A Charlie Brown Christmas \\
&True answer: & A Charlie Brown Christmas \\
\Xhline{1.5pt}
\end{tabularx}}
\caption{Selected samples where the adjusted response (by adding the offset) better aligns with the ground truth compared to the original response (without the offset).}
\label{tab:alleviation}
\end{table}

\end{document}